# Granular Learning with Deep Generative Models using Highly Contaminated Data

*John Just**

[a]*Iowa State University, Ames, IA*

A B S T R A C T

An approach to utilize recent advances in deep generative models for anomaly detection in a granular (continuous) sense on a real-world image dataset with quality issues is detailed using recent normalizing flow models, with implications in many other applications/domains/data types.  The approach is completely unsupervised (no annotations available) but qualitatively shown to provide accurate semantic labeling for images via heatmaps of the scaled log-likelihood overlaid on the images.  When sorted based on the median values per image, clear trends in quality are observed.  Furthermore, downstream classification is shown to be possible and effective via a weakly supervised approach using the log-likelihood output from a normalizing flow model as a training signal for a feature-extracting convolutional neural network.  The pre-linear dense layer outputs on the CNN are shown to disentangle high level representations and efficiently cluster various quality issues.  Thus, an entirely non-annotated (fully unsupervised) approach is shown possible for accurate estimation and classification of quality issues..

## 1. Introduction

Explicit generative models with the capacity to learn the complex probability distributions of high-dimensional data have matured significantly over the last five years in terms of log-likelihood scores on standard public tabular (Dua & Taniskidou, 2017) and image (e.g. MNIST, CIFAR10) datasets.  However, applying these same algorithms to real-world uses such as anomaly detection have not fared as well as the improvements in LL scores, with even a simple full-covariance Gaussian performing much better in some cases due to difficulties with optimizing the networks (Just & Ghosal, 2019) (Hendrycks, Mazeika, & Dietterich, 2019) (Choi, Jan, & Alaxander, 2019) (Nalisnick, Matsukawa, Teh Why, Gorur, & Lakshminarayanan, 2019) (Shafaei, Schmidt, & Little, 2019).  It has been shown that under the right conditions current optimization protocols for training models based on stochastic gradient decent, along with the right data preprocessing and network architectures, can be effective at training models for anomaly detection (Just & Ghosal, 2019).  This is good news since the flexibility of a single full-covariance Gaussian is far less than density models based on neural networks, which have highly scalable modeling capacity and can contort to a wider swath of atypical manifold shapes (De Cao, Aziz, & Titov, 2019) (Dinh, Sohl-Dickstein, & Bengio, 2017) (Papamakarios, Pavlakou, & Murray, 2017).  The potential use of such models extends beyond simply Boolean anomaly detection.  (Just & Ghosal, 2019) in particular demonstrated that the "ones" class from the MNIST digits images are assigned higher probability than other digits when a neural density model is trained on the Fashion MNIST dataset, due to similarity with the trousers class.  This result indicates a sort of granularity to the log likelihood (LL) signal that can be further leveraged, and stimulates the idea that perhaps such a model can learn granular coding of the data by LL even in the presence of a potentially large amount of anomalous data contaminating the dataset.  This stands in contrast to expectations by work such as (Hendrycks, Mazeika, & Dietterich, 2019) which suggest outlier exposure as a means to overcome poor anomaly detection performance in the first place.  If learning the novelty of a data point on a continuous level in the presence of contaminated data is possible and effective, then a sort of continuous quality (or anomaly) estimation signal with a generative model via the proxy of log-likelihood could extend to many applications in the food, drug, medical, military, and agricultural applications, just to name a few.

### 1.1. Normalizing Flows & Deep Generative Models

The primary enabling concepts of this work depend on the use of a recent subclass of deep generative models called normalizing flows (Dinh, Krueger, & Bengio, 2014) (Dinh, Sohl-Dickstein, & Bengio, 2017) (Papamakarios, Pavlakou, & Murray, 2017) (De Cao, Aziz, & Titov, 2019).  These models provide a highly flexible means to parameterize the probability density function (PDF) of complex and high-dimension data.  On a metalevel, deep generative models which are both explicit and tractable currently operate under two separate paradigms from basic probability theory.  The first one of these uses the chain rule of probability that decomposes the joint distribution into the product of conditional distributions.  This is leveraged by models such as pixel CNN++ (Salimans, Karpathy, Chen, & Kingma, 2017) and Masked Autoencoder for Distribution Estimation (MADE) (Germain, Gregor, Murray, & Larochelle, 2015).  The second uses the change of variables technique, typically under the assumption that the final latent space distribution (which is of the same dimensionality as the original) consists of independent random variables.  Thus in a sense it is a form of probabilistic independent component analysis (Dinh, Krueger, & Bengio, 2014) (Hyvarinen, Karhunen, & Oja, 2001).  These have been termed normalizing flows, and have so far matured beyond their conditional model counterparts to achieve best log-likelihood on benchmarks (De Cao, Aziz, & Titov, 2019).

* *Corresponding author.*
  E-mail address: justjo@iastate.edu

*1.2. Unsupervised Disentangled Representation Learning*

In many applications, being able to obtain an accurate estimate of the novelty of a new data point from contaminated data is an end in itself. E.g. it could translate to algorithms for medical devices such as early seizure warning systems that adapt to each individual to maximize accuracy. However, consistent methods for unsupervised disentangled representation learning of human interpretable features from data would make great strides in artificial general intelligence, since machines could then provide useful classifying knowledge regarding new and large data sources back to humans without explicit and tedious instruction. There have been attempts to achieve just this with some reported success on simple examples (Higgins, et al., 2017), but no break throughs since the problem is theoretically ill-posed (Locatello, et al., 2019), and highly dependent on the priors enforced on the solution (e.g. model architecture, causality, hierarchy, etc) as well as the tasks they learn on (Lake, Ullman, Tenenbaum, & Gershman, 2016). At the same time, for more specific applications and tasks there has been some encouraging progress from self-supervised methods at extracting human-interpretable features from convolutional neural networks (CNNs) (Kolesnikov, Zhai, & Beyer, 2019). Also, when a supervised task related to human-interpretable annotations is used to train a network, it has been observed that the extracted features tend to cluster (denote similarity between) images in a similar manner to humans (Zhang, Isola, Efros, Shechtman, & Wang, 2018).

*1.3. Contributions*

A fully unsupervised (no annotations) approach to semantically label images with the learned probability model, show that a continuous signal can be obtained from high-dimensional data sources which can serve as a direct proxy to quality, or simply as a magnitude/level on novelty of a given data point.

1. It is shown possible to obtain reliable and accurate anomaly detection via low log-likelihood scores from normalizing flow models trained on highly contaminated data from high-dimensional data sources (images in this case) is shown. Thus, a dataset of clean (no outliers) is not necessary in practice to learn a good model of the data probability distribution.
2. The results of (1) translates to a fully unsupervised (no annotations) approach to fine-grained semantic labelling of images with a scaled log-likelihood score from the learned probability model, which can serve as a direct proxy to quality, a magnitude/level on novelty of a given data point, or a way to highlight parts of the image for downstream tasks.
3. It is proposed and shown that the "quality" signal (log-likelihood) can be leveraged in a weakly supervised fashion to train a feature extracting classification model. As a result, the pre-linear layer (assuming a two-layer dense neural network is used at the output of a convolutional architecture) provides disentangled high-level features that are useful representative of the features that humans are interested in. In doing so a fully unsupervised approach to disentangled feature extraction is made possible.

The positive results described herein are in contrast to the anomaly detection problems noted in other works (Choi, Jan, & Alaxander, 2019) (Hendrycks, Mazeika, & Dietterich, 2019) (Nalisnick, Matsukawa, Teh Why, Gorur, & Lakshminarayanan, 2019) (Shafaei, Schmidt, & Little, 2019), but leverages the results and learnings of (Just & Ghosal, 2019), including the use of newer models, transforming to a different basis, and avoiding convolutional architectures. The implications for these results are enormous, with the possibility to use such methods with virtually any type of signal or data source (not just images) and could extend to various industries for which quality estimation of identification of novel data is important. Some examples include developing more accurate medical instrumentation (e.g. heart arrhythmia, epilepsy), reflectance spectroscopy for medicine (e.g. carcinogenic substances introduced during manufacturing), food safety (e.g. melamine contamination) and quality (counterfeit and contaminated spices), and identifying crop disease from aerial imagery. The especially novel part of this being that no knowledge of the potential contaminants, or annotated data, is necessary ahead of time. This would reduce the stress of keeping up with the latest problems, and could pass the burden to the algorithms.

## 2. Experimental Setup

*2.1. Data Summary*

The data used to prove the application of concepts and methods detailed in this work consists of images of corn samples harvested over three separate years from typical fields in the Midwest (Table 1). Each image is a different sample as harvested from a grain combine. Such data contains real-world variations in material such as color, size, shape, orientation, and various contaminants such as leaves, twigs, broken/rotten/cob pieces, and chaff. Any material other than clean intact (unbroken) corn kernels is generally referred to as MOG (material other than grain) and represents a quality concern. In harvesting applications, knowledge of quality as is shown in this work presents an opportunity for feedback control and ultimately intelligent harvesting decisions leading to autonomy of the machine. Four separate camera systems were used to collect the images in the datasets. Each system was intended to be identical, but it is expected that tolerances due to variations in the lighting source and the camera hardware exist. The training and validation data were from 2017 & 2018, and the test data from 2015. The original/full-scale size of the images is 460x640x3.

**Table 1:** Relevant meta data related to the datasets used in this work. Year and # cameras indicate the diversity of factors from which the data originates.

| Year | # images | # Cameras |
|------|----------|-----------|
| 2017 | 386      | 2         |
| 2018 | 91       | 1         |
| 2015 | 336      | 1         |

*2.2. BNAF VS MAF*

While Block Neural Autoregressive Flows (BNAF) (De Cao, Aziz, & Titov, 2019) and Masked Autoregressive Flows (MAF) (Papamakarios, Pavlakou, & Murray, 2017) are both normalizing flows, they have significant differences and each their own advantages and drawbacks. As essentially a stack of MADEs formed to increase modeling capacity, MAF takes direct advantage of the fact that MADE (and the probability chain rule) forms a lower triangular matrix dependence structure, and thus the determinant of the Jacobian for each flow, which is required by the change of variables procedure, is simply the product of the diagonal entries. In the case of MAF this is trivial to compute since each flow is typically just an affine transformation of the input random variables (RV), albeit using a scale and bias for each conditioned variable that are a complex function of the conditioning variables. Because the MADE architecture can be thought of as reusing the transformations on the conditioning variables for all downstream conditioned variables, MAF is a relatively efficient parameterization. BNAF on the other hand trades parameter efficiency for flexibility, while still retaining the concept of a lower-triangular dependence structure for ease of calculating the Jacobian determinant. Each variable transformation per flow is an unrestricted dense neural network function of the conditioning variables, and a monotonic neural network with regards to the variable itself. Thus, each variable per flow has a unique transformation, and is by design less parameter efficient than MAF. While flexible, this parameterization makes BNAF difficult to employ even with small images like CIFAR10 (32x32x3) since there is not enough memory available on a typical high-end GPU (~12GB VRAM). In practice this requires some kind of dimensionality reduction, although conveniently it was found that using linear techniques (e.g. singular value decomposition) help with the optimization enough to provide a net gain in anomaly detection, even though some information is thrown away in the process (Just & Ghosal, 2019).

### 2.3. Modeling Approach

In order to train a model that could produce a quality (anomaly) heatmap over the image, the density model was built on smaller crops from the image as depicted in Figure 1. Subsequently the trained model could be swept over the image in the same way a windowed filter would be, and a kind of heatmap of LL produced. Several experiments were performed that trained both BNAF and MAF models and examined performance of the heatmaps qualitatively (since no actual annotation was available). The experiments included image crops from full resolution and reduced (to 25% original size) images as described in Table 2. When using the full-sized image, even though the crop is relatively small compared to objects in the image, at 46x46x3 (6348) the dimensionality is very large for a typical dense-type neural model like the ones used herein. Although MAF could technically be effectively used at this level due to the efficient parameter reuse architecture of MADE that it leverages, as was done with CIFAR10 in (Just & Ghosal, 2019), BNAF did not scale as well. Instead of severely restricting the network architectures such that the entire model can be trained on a typical GPU (12GB VRAM), the procedures of (Just & Ghosal, 2019) were followed to both reduce dimensionality and achieve a potentially better optimized result. The dimensionality is reduced via SVD to 100 components for the full-sized images, but the reduced sized images could be modeled in full dimension (363). This number was not identified as ideal through extensive tuning, but simply worked well enough from previous experience and produced results well enough in this case to prove the concepts in this work. Such factors should be extensively explored and tuned prior to deployment for any application. The LL scores are not published since they are in effect meaningless for the purposes of this work due to known lack of correlation with anomaly detection performance (Just & Ghosal, 2019).

Table 2: Nominal information regarding density modelling. Both BNAF and MAF were used at two different resolution levels. Dimensionality reduction of the full-resolution random crops was necessary for BNAF for the full resolution case, since otherwise the large number of dimensions would make training on a GPU with 12GB VRAM implausible for most architectures, and was also employed for MAF for a fair comparison.

| Model | Architecture | Resolution | size | Dimensions |
|---|---|---|---|---|
| MAF | 5 Flows, 100 relu, Batch Norm | 100% (460x640) | 46x46 (6348) | 100 (SVD) |
| BNAF | 6 Flows, 12 tanh | 100% (460x640) | 46x46 (6348) | 100 (SVD) |
| MAF | 5 Flows, 100 relu, Batch Norm | 25% (115x160) | 11x11 (363) | 363 |
| BNAF | 6 Flows, 12 tanh | 25% (115x160) | 11x11 (363) | 363 |

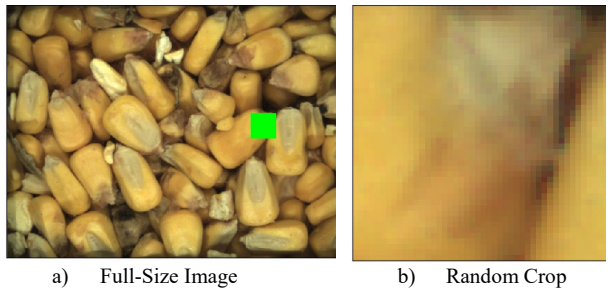

a) Full-Size Image    b) Random Crop

Figure 1: A green 46x46x3 pixel box is shown in the image in (a) where a crop is taken, with the actual cropped image in (b)

Because the training procedure used random crops from the images in the training set, of which a very large number of combinations was possible and duplications of the same crop very uncommon, the validation data for early stopping simply leveraged the same pipeline of images as the training data. This was an efficient use of the data since the quantity of images was not large.

Code for training (In both cases the code has been built for TensorFlow 2.0)
- MAF: https://github.com/johnpjust/MAF_GQ_images_tf20.
- BNAF: https://github.com/johnpjust/BNAF_GQ_images.

### 3. Quality Estimation Results

While it is emphasized that the solution presented to the quality estimation problem is an unsupervised one since no annotations are available, it is also recognized that the qualitative assessment that commences during model selection and tuning is a form of supervision. There is no need to resolve this since it is rare that an algorithm would be deployed without some kind of confidence that it will succeed in the task required of it. Also, very little changes were implemented from (Just & Ghosal, 2019) to obtain the results here, thus it provides confidence this strategy is fairly robust and will work well without much tuning regardless. In order to qualitatively assess the performance, each image was overlaid with a kind of heatmap of the LL (scaled for optimal visualization as an image), and compared with the original in Figure 3 for the training/validation data and Figure 4 for the test data. Between MAF and BNAF and the two resolutions examined, there was relatively high correlation in the LL values, so note that overall the approach is fairly robust against these types of choices. The heatmap were obtained by sweeping the 46x46x3 crop window over the full-resolution image at strides of eight pixels horizontally and vertically and calculating the LL (after reducing dimensionality to 100 components via SVD) at each location using the trained model. Dark and red colors indicate lower LL, and therefore lower quality. The result produced a total of 3570 LL estimates per image using the full-resolution model. Figure 2 shows the box plots of all 3570 LL values for six representative images in the train and test datasets. The images were selected by binning the LL values by the average of the $25^{th}$ and $50^{th}$ percentiles for each image, and taking a representative image from each bin in order to observe the full range in quality found in each dataset. Overall the ranges of LL were very similar for the train and test sets, and resulting quality estimates comparable in each bin for Figure 3 and Figure 4.

Observing the lowest LL bins in the train and test set in Figure 3 & Figure 4 show very different images, but closer inspection shows similar levels of quality due to different quality factors. In the training data the corn is unusually bright yellow, and contains a large amount of broken and small pieces. Conversely, the image for the lowest bin in the test data contains a large amount of trash and immature kernels. The corresponding heatmaps for each are very good but not perfect. There are some instances where quality issues such as leaves or other material other than grain (MOG) are not completely shaded in red. In other cases, some parts of kernels are shaded even though the kernel does not appear to have any obvious quality issues. This may be indicative of potential future improvements by further model & window size tuning and selection. However, these cases are not substantial and the level of the quality is clearly seen to be increasing with the binned values (left to right). Moreover, in some of the false positive cases (identifying low quality when none is observable) the algorithm may be finding non-trivial abnormalities with sizes/shapes/colors that are difficult for a human to observe (i.e., there may be underlying quality factors that aren't as obvious as broken kernels and MOG).

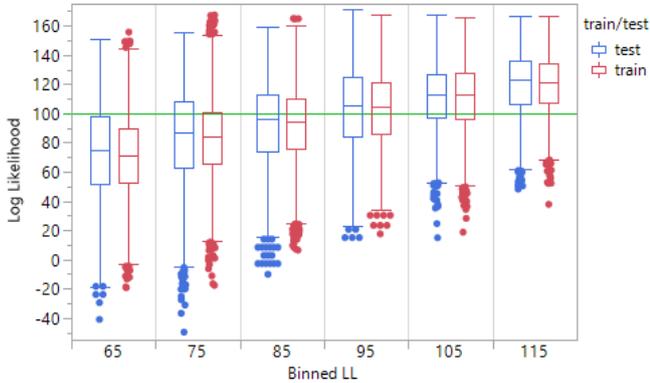

Figure 2: Per-image box plots for representative images from the train and test sets at increments of ten for binned LL levels. The binned LL corresponds to the average of the $25^{th}$ and $50^{th}$ percentile for each image. The corresponding images are shown in Figure 3 for train data and Figure 4 for test data.

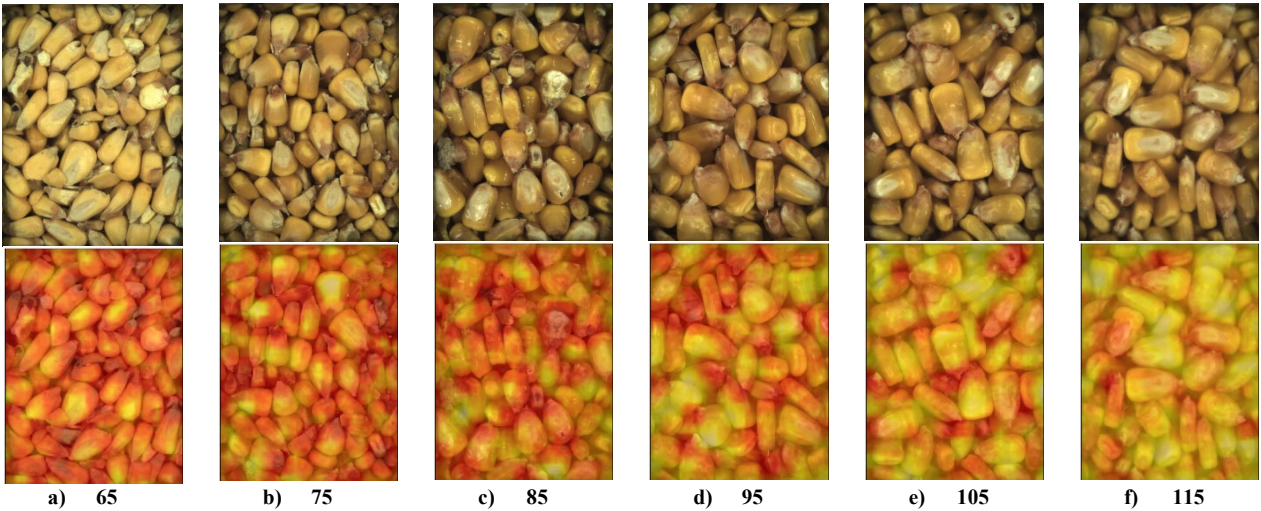

a) 65  b) 75  c) 85  d) 95  e) 105  f) 115

Figure 3: Training Data heat maps for the box plots in Figure 2. The top image is the original and the bottom image in each case is the same image overlaid with a semi-transparent heat map. Low LL (low quality) is denoted by darker/more red shades.

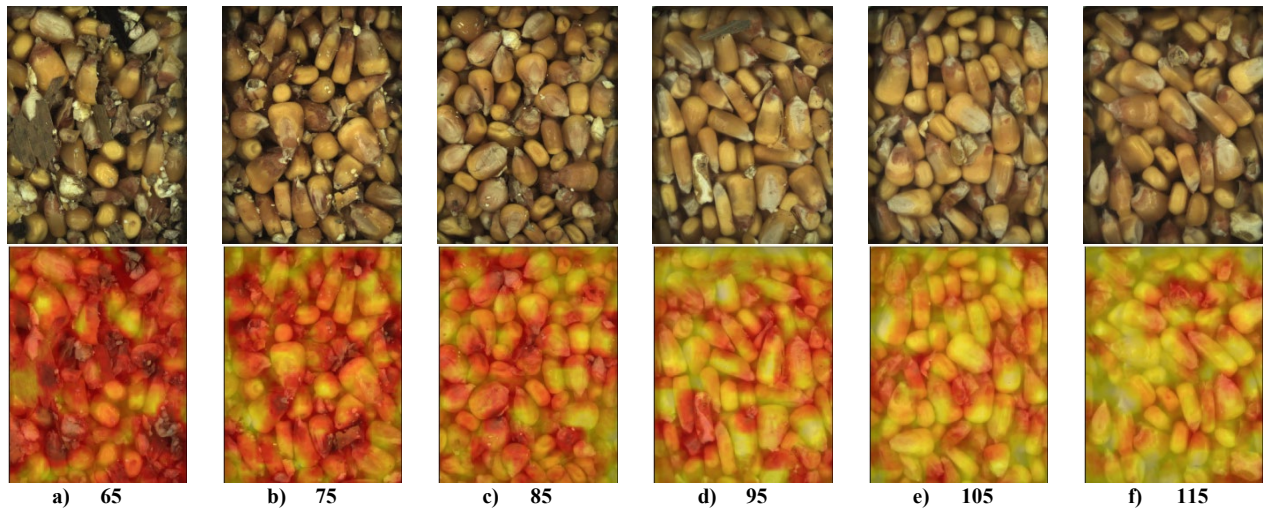

**Figure 4:** Test Data heat maps for the box plots in Figure 2. The top image is the original and the bottom image in each case is the same image overlaid with a semi-transparent heat map. Low LL (low quality) is denoted by darker/more red shades. Some issues like the large brown leaf piece in (a) wasn't completely identified as low-quality with dark red everywhere, which may indicate opportunity to improve results by further modeling tuning and architectural selection.

## 4. Downstream Classification Tasks

The normalizing flow models used herein have shown to be very good density models, but the neural architectures are a highly restricted form such that they are not very conducive to interpretable feature extraction. The downstream task of classifying novel/anomalous data (e.g. poor quality) can be highly useful such as in the example presented in this work, but the density models will not provide it. Instead the results from (Zhang, Isola, Efros, Shechtman, & Wang, 2018) which highlights the effectiveness of convolutional architectures at interpretable features for judging image similarity, inspire the use of typical residual connection CNN networks to do this work. (Kolesnikov, Zhai, & Beyer, 2019) find that the pre-logit layer from classification models works well with residual-connection networks. In that case they find very good results with a slightly modified residual network using fully invertible connections inspired by (Dinh, Sohl-Dickstein, & Bengio, 2017). Although the models explored for this work only uses the more common residual connections, it is noted since it may be a good avenue to explore in future work or in other applications. The key to training the feature extraction network then lies in the target signal. In this case no labels or annotated data exist, but there does exist the quality estimate in the form of a LL from the density model, which has already shown to also correlate with features that are human-interpretable as quality issues. Since this is a regression problem and not classification though, the pre-linear layer is used, which is positioned similarly as the pre-logits layer in the layers hierarchy. To train the model then random crops were taken from the images as they were during the training procedure for the generative model, but in this case a pre-trained generative model estimates the LL from the crop, which is used as the target for the CNN model. Figure 5 shows the relative window/crop size used, which is considerably larger than the one in Figure 1. Larger window sizes were used at this stage primarily so that the visualization produced in Figure 6 would be clearer, whereas the quality estimator was earlier aiming at producing a high resolution heatmap of quality over the image. However, there was a high correlation between the median LL per image produced by smaller and larger windows, which again underscores the robustness and generality of the overall strategy/approach.

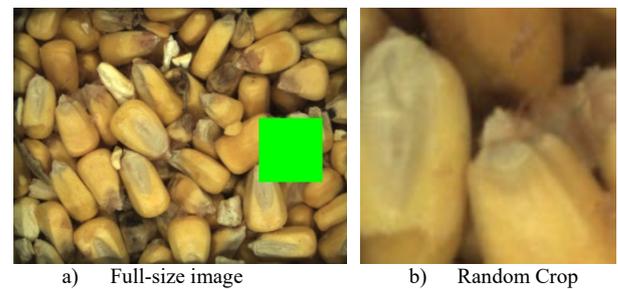

a) Full-size image  b) Random Crop

Figure 5: The random crop location from the full-size image is shown in (a) by a green box. The same random crop is shown in (b). Note the full resolution image is shown here for clarity, but the actual images used in training and evaluation were down sampled to 25% of the original size.

The CNN architecture was inspired mostly by the Resnet V2 architecture (He, Zhang, Ren, & Sun, 2016). The ultimate goal in this part was to obtain a kind of disentangled representation in terms of high-level features that are concerned with quality, and to that end it worked very well as is shown in Figure 6. The key was restricting the number of activations in the pre-linear layer to three units for this particular application, and that is part of the weak supervision required. Using more units than that caused the features to be spread across more units and less interpretable. All three features from that layer have direct meaning corresponding to yellowness, particle size, and whiteness, which are proxies to quality factors like broken pieces, immature kernels and cob pieces, and leaf trash or rotten kernels. When this is combined with the likelihood signal it is possible to discern when a quality issue exists and then it can be classified, since in some cases the extremes of these were still healthy kernels (e.g. large, dark kernels). In this case note that blue colored points are low LL (low quality) and red is high LL (high quality).

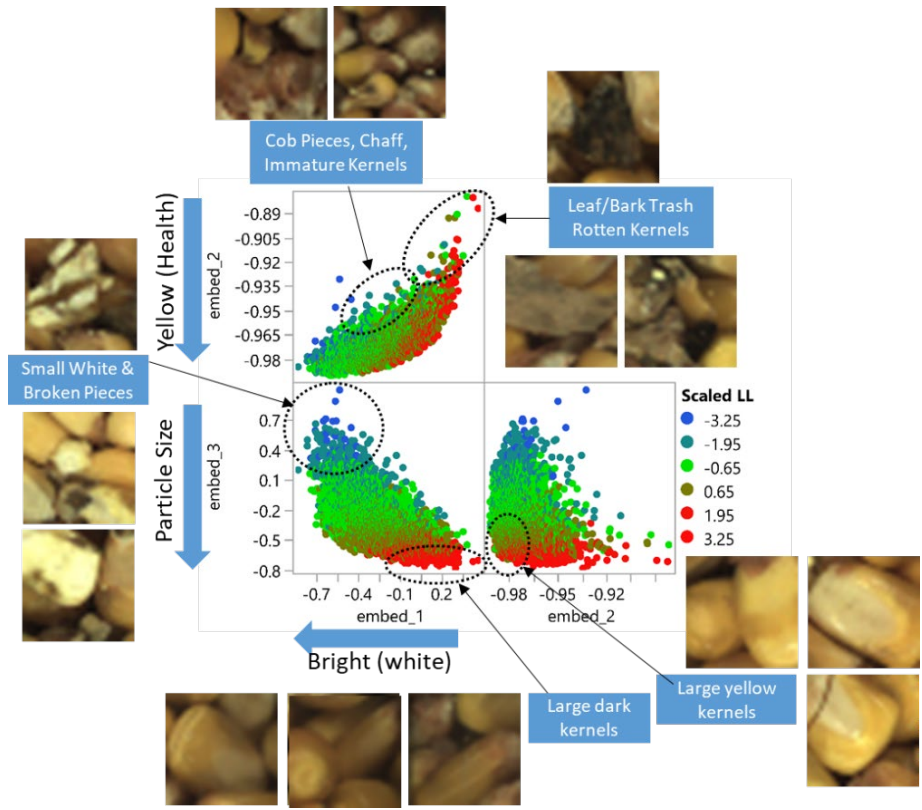

**Figure 6:** A scatterplot of the pre-linear layer feature space of the CNN, colored by the LL score (scaled for visualization). Example crops from certain areas highlighted show the clustering of images based on human-interpretable factors, and the overall disentangled representation achieved in this feature space. Note that data points identified as high quality/LL (red) in the leaf/bark and rotten kernels cluster are just very large dark-colored kernels (as would be expected).

Code for training a feature extractor using a pre-trained MAF density model is available at https://github.com/johnpjust/GQC_featureExtraction.

## 5. Discussion

With implications ranging from the food and drug industry, to medical instrumentation, military, and agricultural applications, it is shown that a fully label-free (unsupervised) approach utilizing artificial intelligence algorithms to estimate novelty and/or classification with high-dimensional data is not only feasible, but can be highly effective in cases where obtaining annotated data would be quite impractical. The methods detailed in this work are not in any way limited to the example shown, but could be easily and readily extended to achieve cutting edge results in applications such as heart or seizure monitoring devices, or detecting food and medicine quality or counterfeit spices with reflectance spectroscopy, or disease monitoring of crops from aerial imagery. In the example presented in this work, the semantic and granular (both spatially and on a continuous scale) labeling of the quality of grain in images was performed in an unsupervised fashion with normalizing flow deep generative models. This involved overlaying a heatmap of the scaled log-lowlihood spatially on the images, and also by utilizing point values for each image to sort by overall quality. Furthermore, it is shown that training a feature extracting convolutional neural network with the output (log-likelihood) of a pre-trained deep generative model results in a disentangled representation in the pre-linear layer, ultimately providing a highly effective unsupervised (or at most weakly supervised) means for disentangled representation learning.

## Acknowledgements

We gratefully acknowledge the support of NVIDIA Corporation with the donation of the Titan V GPUs used for this research.

\* *Corresponding author.*
E-mail address: justjo@iastate.edu